\documentclass[letterpaper, 10 pt, conference]{ieeeconf}  

\usepackage{times}
\usepackage[numbers]{natbib}
\usepackage{multicol}
\usepackage[bookmarks=true]{hyperref}

\usepackage{lipsum}
\usepackage{comment}
\usepackage{balance}
\usepackage{graphicx}
\usepackage[table]{xcolor}
\usepackage{caption}
\usepackage{listings}
\usepackage{amsmath} 
\usepackage{booktabs}
\usepackage{cuted}
\usepackage{xcolor}

\definecolor{codegreen}{rgb}{0,0.6,0}
\definecolor{codegray}{rgb}{0.5,0.5,0.5}
\definecolor{codepurple}{rgb}{0.58,0,0.82}
\definecolor{backcolour}{rgb}{0.95,0.95,0.92}

\lstdefinestyle{mystyle}{
    backgroundcolor=\color{backcolour},
    commentstyle=\color{codegreen},
    keywordstyle=\color{magenta},
    numberstyle=\tiny\color{codegray},
    stringstyle=\color{codepurple},
    basicstyle=\ttfamily,
    breaklines=true,
    captionpos=b,
    keepspaces=true,
    numbers=left,
    numbersep=5pt,
    showstringspaces=false,
    tabsize=2
}
\usepackage{minted}
\usepackage{xcolor}
\definecolor{bg}{HTML}{f2f2ea}
\usepackage{float} 

\usepackage{seqsplit}

\IEEEoverridecommandlockouts                              

\usepackage{tabularx} 
\title{\Large \bf
A New Software Tool for Generating and Visualizing Robot Self-Collision Matrices
}

\author{\authorblockN{Roshan Klein-Seetharaman and Daniel Rakita}
\authorblockA{Department of Computer Science, 
Yale University}
\authorblockA{\{roshan.klein-seetharaman, daniel.rakita\}@yale.edu}
}

\begin{document}

\maketitle
\thispagestyle{empty}
\pagestyle{empty}


\begin{strip}
  \centering
  \includegraphics[width=0.99\textwidth]{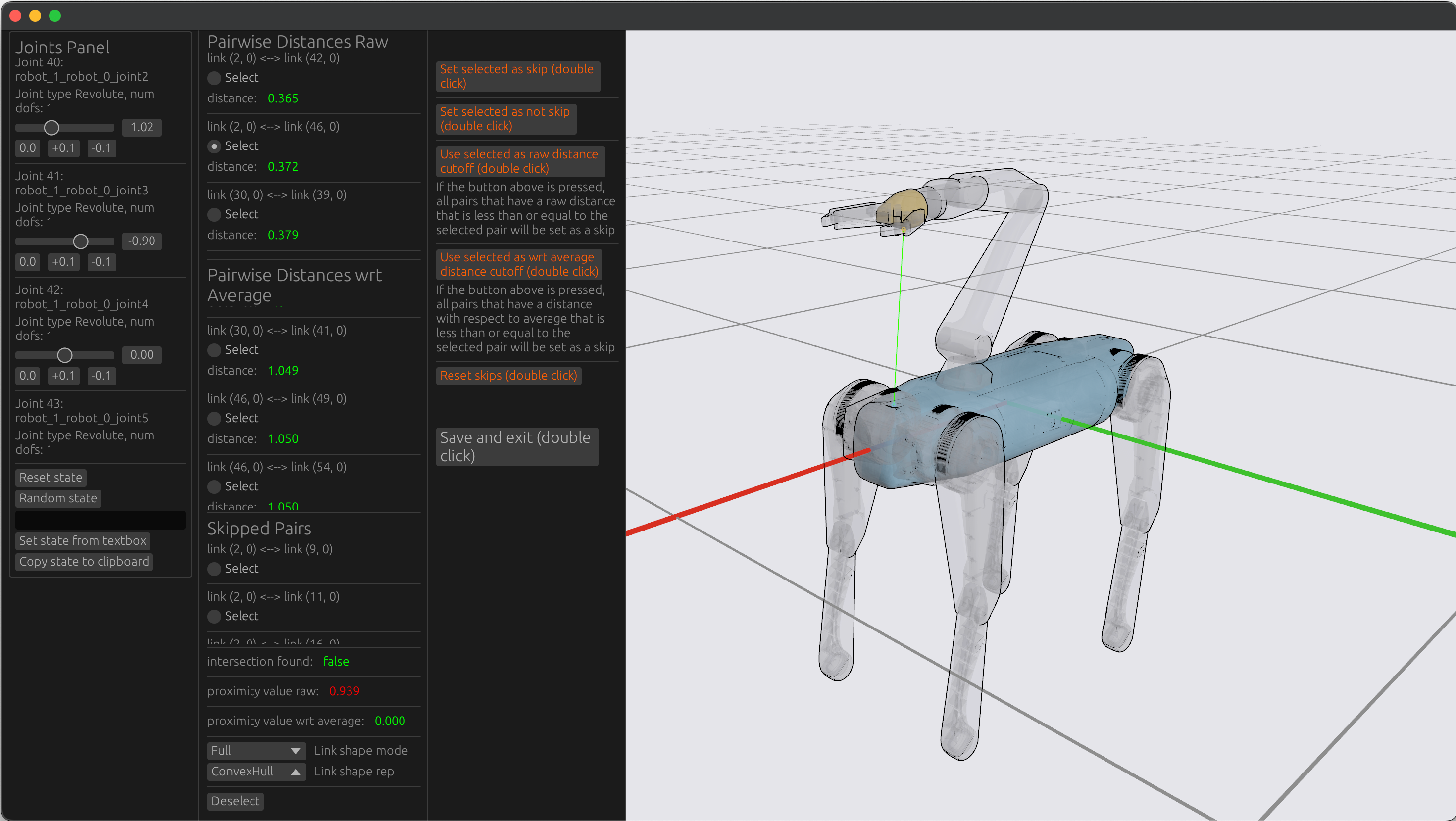}
  \captionof{figure}{Overview of our interactive self-collision matrix tool. Starting from a URDF and associated meshes, the system automatically generates six shape-based representations (spheres, OBBs, convex hulls, and convex decompositions) and infers skip sets across 100k sampled configurations. Users can dynamically explore robot states, inspect proximities, and refine skip decisions in a Bevy-based visualization, with results exported to JSON/YAML for direct use in planning and simulation frameworks. }
  \label{fig:teaser}
  \vspace{-0.3cm}
\end{strip}
\begin{abstract}

In robotics, it is common to check whether a given robot state results in self-intersection (i.e., a self-collision query) or to assess its distance from such an intersection (i.e., a self-proximity query). These checks are typically performed between pairs of shapes attached to different robot links. However, many of these shape pairs can be excluded in advance, as their configurations are known to always or never result in contact. This information is typically encoded in a \textit{self-collision matrix}, where each entry $(i, j)$ indicates whether a check should be performed between shape $i$ and shape $j$.  While the MoveIt Setup Assistant is widely used to generate such matrices, current tools are limited by static visualization, lack of proximity support, rigid single-geometry assumptions, and tedious refinement workflows, hindering flexibility and reuse in downstream robotics applications.  In this work, we introduce an interactive tool that overcomes these limitations by generating and visualizing self-collision matrices across multiple shape representations, enabling dynamic inspection, filtering, and refinement of shape pairs. Outputs are provided in both JSON and YAML for easy integration. The system is implemented in Rust and uses the Bevy game engine to deliver high-quality visualizations. We demonstrate its effectiveness on multiple robot platforms, showing that matrices generated using diverse shape types yield faster and more accurate self-collision and self-proximity queries.      

\end{abstract}

\section{Introduction}
\label{sec:introduction}

In robotics, it is common to determine whether a given robot state produces an intersection with itself (a self-collision query) or, alternatively, to evaluate its distance from such an intersection (a self-proximity query).  Self-collision and self-proximity queries are commonly conducted by iterating through pairs of shapes attached to some robot link and either checking for intersections or distances between these shapes, respectively.

In practice, many shape pairs are excluded from these checks to improve efficiency, since it can be determined in advance that they do not influence the result.  For instance, shape pairs may be excluded if it is determined that they never collide, are perpetually in collision, or belong to the same rigid link.  A common way to store this information is in a self-collision matrix, where each entry ($i$, $j$) indicates whether a collision or proximity check should be carried out between shape $i$ and shape $j$.

In many robotics workflows, the MoveIt Setup Assistant has become the standard tool for generating self-collision matrices.  The MoveIt Setup Assistant generates a self-collision matrix by sampling robot configurations to identify link pairs that need not be checked, and stores the resulting specification in a YAML file within the MoveIt configuration package. 

While this tool has served a useful role in the robotics community for several years, some limitations remain: (1) The RViz visualization in the Setup Assistant is static and does not allow exploration of joint space to verify how link-skip settings affect collision outcomes; (2) The visualization does not expose distances between link shapes, reflecting the Setup Assistant’s focus on discrete self-collision checks rather than continuous proximity queries; (3) Each link can be associated with only a single collision geometry, which must be specified in the URDF in advance. It is not possible, for example, to assign a link multiple geometries corresponding to a convex decomposition; (4) The Setup Assistant produces only one self-collision matrix, tied to the single collision representation defined in the URDF; and (5) Refining this matrix after its creation is cumbersome: if unexpected collision behavior is observed, users must either manually edit the YAML specification or regenerate the matrix entirely.

In this paper, we present a new highly interactive and visually clear software tool for creating self-collision matrices for robot platforms.  The system is implemented in Rust and leverages the Bevy game engine to deliver high-quality visualizations.  At a high level, the tool was designed to address the limitations of the self-collision matrix component in the MoveIt Setup Assistant, as outlined above.  

A core design principle of our tool is support for multiple shape representations, spanning from low to high detail.  To achieve this goal, the system first computes approximating shapes around the given meshes.  Specifically, it automatically creates convex hulls and convex decompositions for each link shape.  Building on these representations, our system supports six shape types: (1) bounding spheres for entire links, (2) OBBs for entire links, (3) convex hulls for entire links, (4) bounding spheres for each convex decomposition sub-component, (5) OBBs for each convex decomposition sub-component, and (6) the convex decomposition sub-components themselves.  Importantly, our system supports separate self-collision matrices for all six shape types, stored in separate JSON and YAML files for easy parsing.     

Next, the Bevy-based visualization environment is fully dynamic, enabling users to clearly observe the effects of the current skip information across different robot joint states.  This information covers not only intersections but also proximities.  Distance data for shape pairs are presented in a scrollable window, sorted in ascending order to highlight which pairs are closest to collision.  Individual shape pairs can be isolated for inspection, with surrounding shapes fading to a low opacity level, leaving only the pertinent shape pair and a line segment connecting their closest points.  Shape pairs can be filtered individually from the list or in bulk using a distance threshold.  The different shape types can be selected from a drop-down menu, allowing this dynamic, interactive process to be applied for each independently.

After the self-collision matrices are established and initial JSON and YAML files are outputted, these matrices can be easily refined going forward.  A single line of code can bring up this interactive visualization environment again for any robot, letting the user further filter shape pairs for any shape type.

In \S\ref{sec:evaluation}, we demonstrate the efficacy of our work through example self-collision matrices generated for several robot platforms.  We show that the self-collision matrices created for the different shape types indeed result in faster and more accurate self intersection and self proximity queries over multiple robots.  We conclude with a discussion on the implications and limitations of our work.  We provide open source code for our software tool.\footnote{\href{https://github.com/Apollo-Lab-Yale/apollo-rust}{https://github.com/Apollo-Lab-Yale/apollo-rust}}  

\section{Related Works}
\label{sec:related_works}

Collision detection and proximity queries are central to robotic motion planning and simulation. Their high computational cost, especially for articulated robots, has motivated extensive work on efficient representations and acceleration methods.

\subsection{Geometric Representations and Bounding Volumes}

Geometric representation strongly affects both accuracy and efficiency. \citet{Hubbard1996ApproximatingPWY} introduced spherical approximations as a fast but coarse method, later extended with bounding volume hierarchies. OBBs \citep{Gottschalk1996OBBTreeAHU} and k-DOPs \citep{Klosowski1998EfficientCDAI} provided tighter fits with tunable trade-offs, while AABB trees \citep{Bergen1997EfficientCDAM} remain foundational for hierarchical methods. GPU-accelerated mesh distance computation \citep{Fan2024gDistEDX} highlights hardware-driven improvements.

Our work differs by unifying six shape types in one framework, enabling systematic comparison across representations. Rather than optimizing collision checks themselves, we focus on deciding which checks to perform, achieving efficiency via pruning.

\subsection{Distance Computation and Proximity Queries}

Distance computation underlies collision avoidance and proximity planning. The GJK algorithm \citep{Gilbert1988AFPAE} remains a cornerstone, with subsequent extensions using swept volumes \citep{Larsen2000FastDQW}, gradient proximity measures \citep{Escande2007ContinuousGPAG}, and octree hierarchies \citep{Jung1996OctreebasedHDAH}. Proxima \citep{rakita2022proxima} formalized time- and accuracy-bounded queries for flexible efficiency-accuracy trade-offs.

We extend this line or research by leveraging proximity not only for faster queries but also for deciding which queries are necessary. Our tool computes distance statistics across sampled configurations and visualizes proximity relationships interactively.

\subsection{Motion Planning and Collision-Aware Optimization}

Collision checking is a critical bottleneck in motion planning, as planners must repeatedly verify the feasibility of sampled states and connecting paths. Algorithms such as RRT~\cite{lavalle2001randomized}, PRM~\cite{kavraki2002probabilistic}, and EST~\cite{hsu1997path} established the foundation of sampling-based planning, but each relies on thousands to millions of collision queries per problem instance. The efficiency of these algorithms is therefore tightly coupled to the performance of the underlying collision detection pipeline. Beyond sampling-based methods, optimization-based approaches such as sequential convex optimization \citep{Schulman2014MotionPWAB} further integrate geometric approximations into continuous planning, highlighting the deep connection between collision checking and overall planner scalability.

Our contribution operates as a preprocessing layer: generating self-collision matrices that any planner can use for immediate efficiency gains. By supporting both coarse and high-fidelity representations, we enable planners to balance feasibility checks with precise validation.

\subsection{Software Infrastructure and Libraries}

Modern robotics frameworks already provide powerful engines for dynamics, planning, and collision detection, including Drake~\cite{drake}, Pinocchio~\citep{Carpentier2019ThePCAJ}, and OMPL~\citep{sucan2012open}. General-purpose libraries such as FCL~\citep{Pan2012FCLAGS} serve as common backends in these systems, while related efforts analyze robot description formats~\citep{Tola2023UnderstandingUAAL} or integrate reasoning with geometry~\citep{Gaschler2018KABouMKAH}. 

Our contribution is complementary: rather than replacing existing libraries, we generate self-collision matrices that can be directly loaded into them. This enables immediate efficiency gains with no modifications to the planning or simulation stack. By ensuring full URDF compatibility and exporting in formats consumed by standard toolchains, our system can seamlessly incorporate into existing workflows. For example, the output of our tool is provided in the Universal Robot Description Directory (URDD) format~\cite{klein2025urdd}.

\section{System Overview}
\label{sec:system_overview}

In this section, we provide a high-level overview of the proposed system. We outline the main components of the pipeline, from input specification to output generation, and describe how these pieces connect to form an integrated workflow. The goal is to convey the structure and flow of the system before examining individual components in detail.

\subsection{Inputs and Outputs}
\label{sec:io}

The primary input to the system is a robot description, provided as a URDF file together with its associated mesh geometries. From these specifications, the system generates a set of intermediate shape representations (described below) that form the basis for constructing self-collision matrices. 

The outputs consist of six pairs of JSON and YAML files, one pair for each supported shape type. Each file encodes a self-collision matrix as a tabular structure, where entry $(i,j)$ indicates whether a collision or proximity check should be performed between the $i$-th and $j$-th shapes. These outputs can be readily consumed by downstream motion planning and simulation frameworks, enabling them to bypass redundant checks and improve efficiency.


\subsection{System Flow}

The overall workflow of the system proceeds through the following sequence of steps:

\begin{enumerate}
    \item The user provides a robot description in the form of a URDF file along with the associated mesh geometries.
    \item The system automatically preprocesses each mesh geometry to generate a set of intermediate shape representations and statistics.
    \item Using the shape representations from the previous step, the system samples 100,000 robot configurations and infers which link pairs can be skipped for each shape type (e.g., pairs that are always in collision, never in collision, or redundant).
    \item An interactive graphical environment is launched, allowing the user to refine the automatically inferred skip sets. This interface provides detailed inspection and adjustment tools to validate or override the system's initial guesses.
    \item After refinement, the system outputs six pairs of YAML and JSON files, one for each supported shape type, specifying which link pairs should be skipped during collision or proximity checks.
    \item At any point, the user can reload the outputs from Step~5 into the interactive environment for further refinement. This functionality can be invoked with a single line of code or a single command from the command line.
\end{enumerate}

This flow integrates automatic pre-processing and inference with an interactive refinement loop, ensuring that the resulting self-collision matrices are both efficient and user-validated.

\subsection{Intermediate Representations and Statistics}
\label{sec:representations}

To support multiple levels of geometric detail, the system automatically computes a range of intermediate shape representations for each robot link. These representations are generated using the \texttt{parry} geometry library in the Rust programming language, allowing them to be computed efficiently and robustly from the input meshes. Specifically, for each link, the system produces:

\begin{figure*}[t!]
\centering
\includegraphics[width=\textwidth]{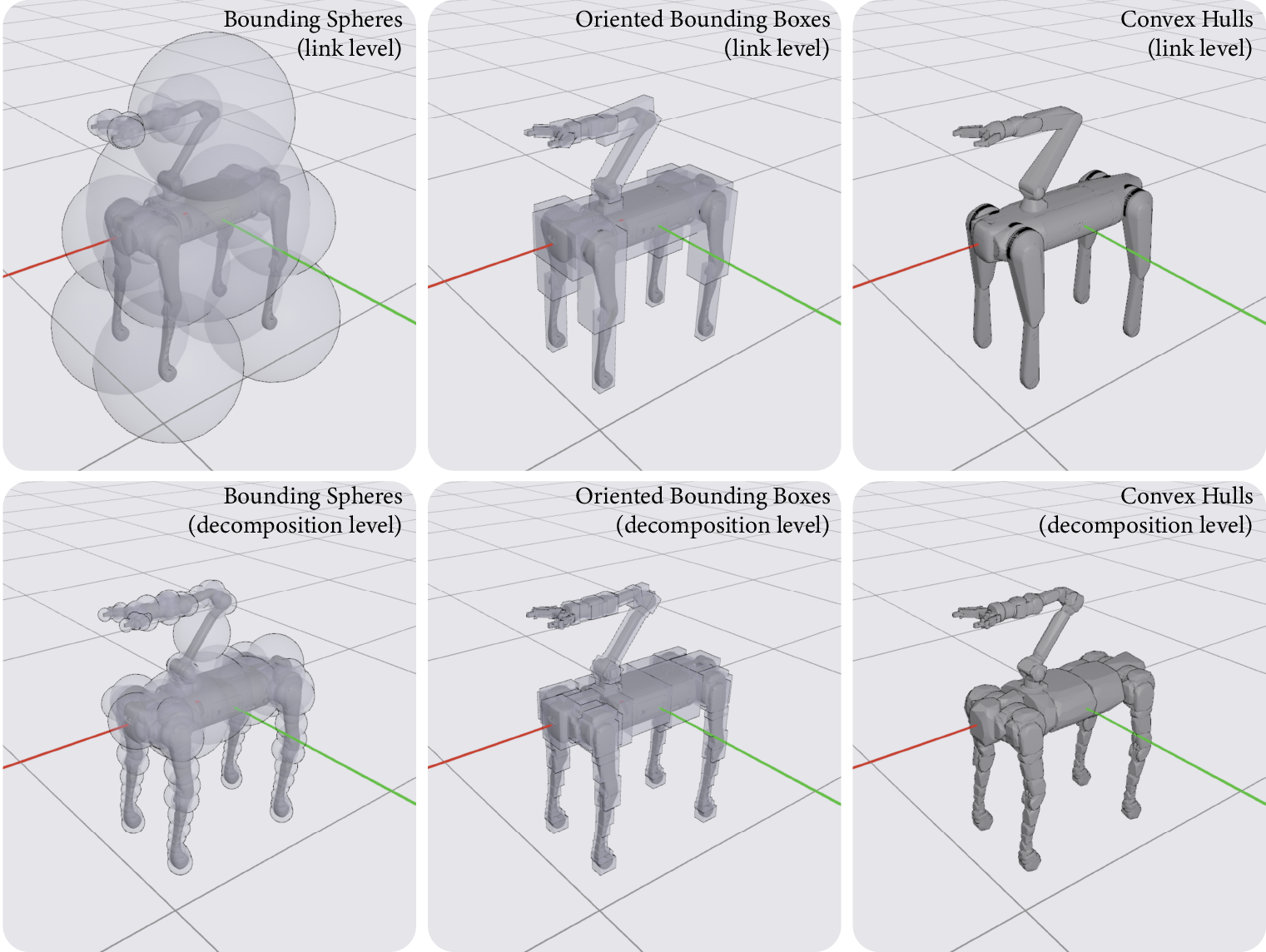}
\caption{Intermediate shape representations generated by our system for each robot link. For every mesh, the tool computes a convex hull, convex decomposition, bounding sphere, and oriented bounding box, along with corresponding decomposed spheres and OBBs. These six representations provide a spectrum of trade-offs between geometric fidelity and computational cost, forming the basis for constructing separate self-collision matrices}
\label{fig:shape_types}
\vspace{-0.4cm}
\end{figure*}

\begin{itemize}
    \item A convex hull enclosing the entire link mesh.
    \item A convex decomposition of the link mesh into multiple convex sub-components.
    \item A bounding sphere enclosing the entire link mesh.
    \item An oriented bounding box (OBB) enclosing the entire link mesh.
    \item A bounding sphere for each convex decomposition sub-component.
    \item An oriented bounding box (OBB) for each convex decomposition sub-component.
\end{itemize}

Together, these six intermediate representations capture a spectrum of trade-offs between geometric accuracy and computational efficiency. They form the basis for constructing the six corresponding self-collision matrices described in the following subsection.

In addition to building these representations, the system collects statistics over the 100{,}000 randomly sampled robot configurations. For each shape type and each pair of shapes $(i,j)$, the system computes the minimum, maximum, and mean observed distances across all samples. We denote these values as $d^{\min}_{i,j}$, $\quad d^{\max}_{i,j}$ and, $\quad d^{\text{mean}}_{i,j}$, respectively. These statistics provide additional insight into the proximity relationships between shape pairs, helping to identify links that are perpetually in collision, never in collision, or typically well-separated.

\subsection{Shape Types}

The system maintains six distinct self-collision matrices, one for each of the intermediate shape representations described in \S\ref{sec:representations}. Each shape type offers a different trade-off between geometric fidelity and computational efficiency, and users can select the most appropriate representation depending on their application. The supported shape types are:

\begin{enumerate}
    \item Bounding spheres (link level): A single bounding sphere enclosing the entire link mesh. This representation is computationally inexpensive but coarse.
    \item Oriented bounding boxes (link level): A single OBB enclosing the entire link mesh, providing a tighter fit than a bounding sphere while remaining efficient.
    \item Convex hulls (link level): A convex hull computed over the entire link mesh, offering higher fidelity at the cost of increased computation.
    \item Bounding spheres (decomposition level): A bounding sphere for each convex decomposition sub-component, yielding a finer-grained but still efficient approximation.
    \item Oriented bounding boxes (decomposition level): An OBB for each convex decomposition sub-component, providing tighter per-part approximations while balancing efficiency.
    \item Convex hulls (decomposition level): The decomposition convex sub-components themselves, giving the highest fidelity representation at the greatest computational cost.
\end{enumerate}

By storing separate self-collision matrices for all six shape types, the system enables downstream frameworks to choose between fast approximations and detailed checks. For instance, bounding spheres can be used for quick feasibility checks, while convex decomposition components provide the most accurate proximity estimates. This flexibility allows the tool to adapt to diverse planning, optimization, and simulation contexts.

\subsection{Interactive Visualization Tool}

A central component of the system is the interactive visualization environment, which enables users to explore, refine, and validate the automatically generated self-collision matrices. The environment is implemented in the Bevy game engine, providing real-time rendering, intuitive interaction, and cross-platform performance within the Rust ecosystem.

The interface includes camera controls, a grid floor for spatial reference, and slider widgets for adjusting each of the robot’s degrees of freedom. A \emph{random configuration} button is also available, allowing the user to quickly sample and visualize a new robot state. These controls make it straightforward to examine how different joint configurations affect potential collisions or proximities.

Users can select among the six supported shape types via a drop-down menu, after which all subsequent interactions apply to the chosen representation. A scrollable window lists all shape pairs associated with the selected type, sorted by their recorded proximity statistics. Selecting a pair from this list highlights the shapes in contrasting colors (blue and yellow), while all other shapes fade to a low opacity with shader-based outlines. This focused visualization helps users easily isolate and reason about the behavior of a given pair.

Matrix refinement is supported through both manual and bulk operations. A highlighted pair can be directly marked as ``skip,'' removing it from further collision or proximity checks. For more efficient editing, the user may also apply bulk rules. For example, the system can mark as skip all pairs whose raw distance $d_{i,j}$ is smaller than that of the currently selected pair. Alternatively, a normalized criterion can be applied using the ratio $\frac{d_{i,j}}{d^{\text{mean}}_{i,j}}$,
skipping all pairs whose normalized distance is below the current threshold. These mechanisms allow users to combine precise pair-level inspection with broad statistical pruning.

Any refinements applied in the visualization environment are immediately reflected in the active self-collision matrices. Updated results can be re-exported on demand as JSON and YAML files, ensuring consistency with the system outputs described in \S\ref{sec:io}. Moreover, the visualization can be relaunched at any time from a single line of code or a simple command-line call, allowing iterative refinement throughout the development process.

In summary, the interactive tool complements the automated preprocessing and sampling pipeline by providing an efficient, user-friendly interface for validation and correction. Built on Bevy’s modern game engine framework, it delivers dynamic visualization, flexible interaction, and seamless integration with downstream robotics workflows.

\section{Evaluation}
\label{sec:evaluation}

\begin{table*}[t!]
\centering
\caption{Ratio of computation time saved for collision checking using the six shape types.  Lower values indicate greater savings achieved through skips in the self-collision matrix.  Range values denote standard deviation.}
\label{tab:collision_speeds}
\resizebox{\textwidth}{!}{%
\begin{tabular}{lcccccc}
\toprule
Robot & Sphere (link) & OBB (link) & Hull (link) & Sphere (decomp) & OBB (decomp) & Hull (decomp) \\
\midrule
UR5   & 0.35 $\pm$ 0.03 & 0.45 $\pm$ 0.04 & 0.55 $\pm$ 0.05 & 0.61 $\pm$ 0.02 & 0.66 $\pm$ 0.03 & 0.76 $\pm$ 0.04 \\
XArm7 & 0.40 $\pm$ 0.02 & 0.38 $\pm$ 0.03 & 0.53 $\pm$ 0.04 & 0.77 $\pm$ 0.05 & 0.68 $\pm$ 0.03 & 0.77 $\pm$ 0.02 \\
B1    & 0.48 $\pm$ 0.03 & 0.36 $\pm$ 0.04 & 0.57 $\pm$ 0.05 & 0.82 $\pm$ 0.02 & 0.84 $\pm$ 0.03 & 0.84 $\pm$ 0.04 \\
\bottomrule
\end{tabular}
}
\end{table*}

\begin{table*}[t!]
\centering
\caption{Ratio of computation time saved for proximity checking using the six shape types.  Lower values indicate greater savings achieved through skips in the self-collision matrix. Range values denote standard deviation.}
\label{tab:proximity_speeds}
\resizebox{\textwidth}{!}{%
\begin{tabular}{lcccccc}
\toprule
Robot & Sphere (link) & OBB (link) & Hull (link) & Sphere (decomp) & OBB (decomp) & Hull (decomp) \\
\midrule
UR5   & 0.45 $\pm$ 0.03 & 0.51 $\pm$ 0.04 & 0.65 $\pm$ 0.05 & 0.78 $\pm$ 0.02 & 0.80 $\pm$ 0.03 & 0.82 $\pm$ 0.04 \\
XArm7 & 0.51 $\pm$ 0.02 & 0.64 $\pm$ 0.05 & 0.65 $\pm$ 0.03 & 0.81 $\pm$ 0.04 & 0.85 $\pm$ 0.02 & 0.79 $\pm$ 0.03 \\
B1    & 0.58 $\pm$ 0.04 & 0.69 $\pm$ 0.03 & 0.78 $\pm$ 0.05 & 0.80 $\pm$ 0.02 & 0.82 $\pm$ 0.04 & 0.80 $\pm$ 0.03 \\
\bottomrule
\end{tabular}
}
\end{table*}

\begin{table*}[t!]
\centering
\caption{Accuracy of skip matrices, reported as the ratio of correctly classified shape pairs.}
\label{tab:matrix_accuracy}
\setlength{\tabcolsep}{6pt}
\resizebox{\textwidth}{!}{%
\begin{tabular}{lcccccc}
\toprule
\textbf{Robot} & Sphere (link) & OBB (link) & Hull (link) & Sphere (decomp) & OBB (decomp) & Hull (decomp) \\
\midrule
UR5    & 44/100 & 85/100 & 100/100 & 100/100 & 100/100 & 100/100 \\
XArm7  & 45/100 & 91/100 & 100/100 & 100/100 & 100/100 & 100/100 \\
B1    & 49/100 & 93/100 & 100/100 & 100/100 & 100/100 & 100/100 \\
\bottomrule
\end{tabular}
}
\end{table*}

We evaluate our system through two experiments. First, we benchmark the six supported shape types to confirm they behave as expected when used for collision and proximity queries. Second, we provide a case study comparing our tool with the MoveIt Setup Assistant, showing how interactive refinement can affect the resulting self-collision matrices.

\subsection{Collision and Proximity Benchmarking}

To assess the efficacy of our proposed system, we tested the generated self-collision matrices on three representative robots. All experiments were conducted on a MacBook Air laptop equipped with an Apple M3 processor and 32GB of RAM. Collision and proximity subroutines were implemented in the Rust programming language using the \texttt{parry} geometry library, with robot models constructed directly in this framework. 

The goal of this evaluation is twofold: (1) to quantify the computational savings provided by the skip information afforded by our system; and (2) to characterize how these savings interact with the accuracy profile of different shape types.


We run this process on three simulated robot platforms: (1) A 6 DOF Universal Robots UR5e\footnote{\href{https://www.universal-robots.com/products/ur5e/}{https://www.universal-robots.com/products/ur5e/}}; (2) A 7 DOF UFactory XArm7\footnote{\href{https://www.ufactory.us/xarm}{https://www.ufactory.us/xarm}}; and (3) A 12 DOF Unitree B1\footnote{\href{https://shop.unitree.com/products/unitree-b1}{https://shop.unitree.com/products/unitree-b1}}.

\subsubsection{Collision checking}  
For each robot and each of the six supported shape types, we measured the average runtime of collision queries across 100{,}000 random configurations. These results were compared to a baseline that performed exhaustive checks with no skips. As reported in Table~\ref{tab:collision_speeds}, we present results as the ratio of runtime with skips enabled to runtime without skips. Values less than 1.0 therefore indicate a reduction in cost, with smaller values reflecting greater savings. 

Across all robots and shape types, this ratio was consistently well below 1.0, showing that the skip matrices accelerate collision queries in every case. For example, link-level sphere approximations yielded ratios near 0.35--0.45, while higher-fidelity decomposed representations achieved ratios closer to 0.75--0.85, still representing substantial gains.  

\subsubsection{Proximity checking}  
We repeated this analysis for proximity queries, where each configuration was tested to compute the minimum distance between all relevant pairs of shapes. Table~\ref{tab:proximity_speeds} again reports the ratio of runtime with skips to the runtime without skips. As with collision queries, all ratios fell below 1.0, confirming consistent computational savings. Coarse link-level approximations produced ratios around 0.45--0.60, while decomposed OBB and convex hull models achieved ratios in the 0.75--0.85 range. These findings demonstrate that skip matrices not only speed up binary collision checks but also extend effectively to continuous proximity queries.

\subsubsection{Accuracy profile of self-collision matrices}  
Finally, we assessed how the inclusion of skips affects accuracy. For each robot and shape type, we visually inspected a set of configurations to verify whether the skip matrices yielded the expected collision outcomes. Table~\ref{tab:matrix_accuracy} reports the ratio of correctly classified shape pairs.

These results highlight a consistent trend: accuracy depends strongly on the fidelity of the geometric representation. Higher-fidelity types (e.g., convex decomposition) achieve near-perfect accuracy, while coarser types (e.g., link-level bounding spheres) remain limited by their geometric simplicity. This evaluation confirms that our system not only reduces computation but also maintains high classification accuracy, giving users the flexibility to choose the representation that best balances efficiency and precision for their application.

\subsection{Case study comparison with MoveIt Setup Assistant}

To illustrate the practical benefits of our system, we conducted a focused case study against the widely used MoveIt Setup Assistant. Because MoveIt supports only a single collision geometry per link, we restrict this comparison to the \emph{convex hull per link} representation, which both systems handle. The study centers on the Orca hand gripper\footnote{\href{https://www.orcahand.com/}{https://www.orcahand.com/}}, a complex multi-fingered mechanism.

\begin{figure}[t!]
\centering
\includegraphics[width=\columnwidth]{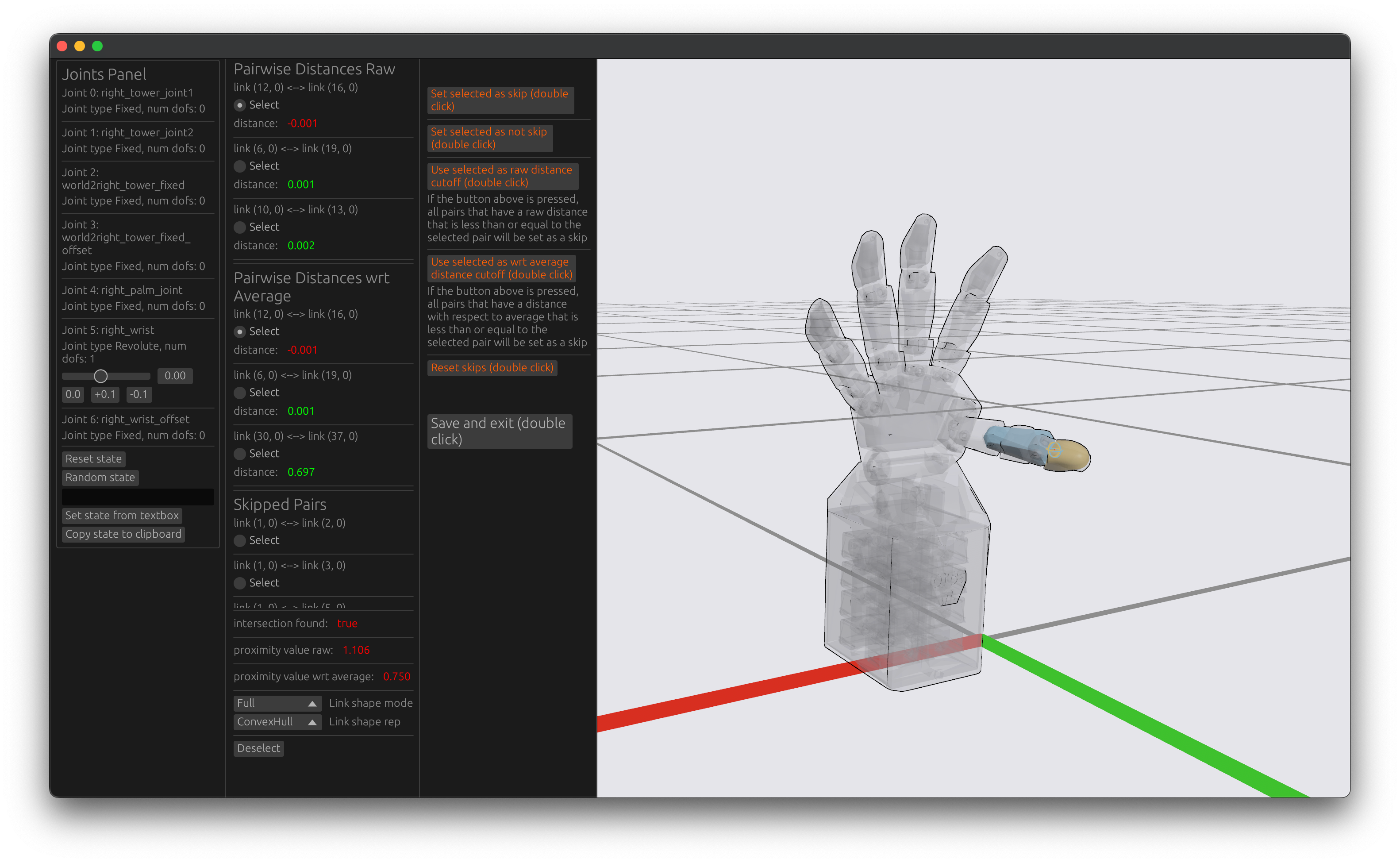}
\caption{Case study on the Orca hand gripper comparing results from the MoveIt Setup Assistant and our interactive refinement tool.}
\label{fig:case_studies}
\vspace{-0.6cm}
\end{figure}

\subsubsection{Methodology}  
We first generated a self-collision matrix using the MoveIt Setup Assistant’s default sampling-based procedure. The resulting YAML file was converted into a format compatible with our system and loaded as the initial specification. From this baseline, we examined shape pairs in our interactive environment using both proximity statistics and visual inspection to identify discrepancies—specifically, pairs that MoveIt left active despite being strong candidates for skips.

\subsubsection{Results}  
Our analysis revealed a clear case of over-conservatism in the MoveIt output: links 12 and 16 were not marked as a skip. These links are not adjacent and, when viewed in our tool, it is immediately apparent that they cannot physically collide under any valid motion. Figure~\ref{fig:case_studies} highlights this scenario: the two links are isolated in contrasting colors while the remainder of the robot fades to low opacity. 


By interactively marking this pair as a skip, our system eliminates redundant checks that MoveIt retained. In practice, MoveIt's sampling-based method is susceptible to such omissions, since rare or edge-case conditions may be under-sampled. Our visualization-driven refinement closes this gap by surfacing these situations clearly, allowing users to make confident corrections with minimal effort.
\section{Discussion}

The results presented in \S\ref{sec:system_overview} highlight several important takeaways about the proposed system. Broadly, the experiments confirm that the generated self-collision matrices provide substantial computational savings, while the case study comparison illustrates how our interactive refinement tools complement existing workflows such as the MoveIt Setup Assistant. Below, we discuss the broader implications, current limitations, and future directions suggested by this work.

\subsection{Implications for Robotics Workflows}
Self-collision matrices are a critical piece of infrastructure in many robotics applications, particularly motion planning, simulation, and optimization. Our evaluation shows that the ability to automatically generate matrices across multiple shape types can yield both efficiency and accuracy benefits. In practice, this means that users no longer face a rigid trade-off between fidelity and performance: coarser shape types (e.g., link-level bounding spheres) can be used for lightweight feasibility checks, while higher-fidelity types (e.g., convex decompositions) provide more accurate proximity estimates where needed. The inclusion of skips further improves performance across all types, enabling fast queries without sacrificing correctness. 

Equally important is the ability to refine matrices interactively. By visualizing robot configurations dynamically and inspecting proximity statistics, users gain insight into how particular shape pairs contribute to collision outcomes. This transparency addresses a long-standing limitation of existing tools, which typically operate as black boxes that output static YAML specifications. The ability to re-import MoveIt’s matrices (§IV.B) and improve them directly in our system demonstrates that our tool is not a replacement, but rather a natural extension that improves downstream robustness.

\subsection{Complementarity with Existing Tools}
The case study comparison underscores the complementary role of our system relative to the MoveIt Setup Assistant. Since MoveIt relies on sampling, its skip matrices are inherently approximate; occasional missed skips or overly conservative decisions are expected. Our tool makes these approximations explicit, enabling users to detect and refine them. For example, both case studies showed shape pairs that should have been marked as skips but were retained in the MoveIt outputs. In our visualization environment, such pairs can be quickly identified through statistics ($d^{\min}_{i,j}$ consistently large) and visual inspection, and then reclassified with minimal effort. This workflow highlights a synergistic relationship: MoveIt provides a fast baseline, and our system supplies the mechanisms for refinement and extension.

\subsection{Limitations}
While promising, the current system has several limitations. First, the scalability of the approach has not yet been fully tested on extremely large or highly articulated robots. Although the six supported shape types span a useful range of fidelity, certain applications may require even richer geometric representations (e.g., mesh-based continuous collision detection). Second, while the interactive environment lowers the burden of refinement, it still requires user input and judgment. Automating more of the refinement process—for instance, by incorporating heuristics that flag likely redundant checks—would further reduce the need for manual inspection. Finally, although we demonstrated compatibility with MoveIt outputs, broader integration with other robotics software ecosystems (e.g., MuJoCo, Drake) would enhance adoption.

\subsection{Future Directions}
Several avenues for future work arise naturally from this study. On the algorithmic side, one direction is to incorporate adaptive sampling strategies that better focus on the regions of configuration space most relevant for planning, reducing the need for brute-force random sampling. Another is to extend the tool to support continuous-time verification, ensuring that pairs marked as skips remain valid not just at sampled configurations but along motion trajectories. From a systems perspective, integrating the tool more directly into planning frameworks could provide real-time feedback during plan execution, where matrices might be refined online as new evidence becomes available. Finally, a promising extension is to couple self-collision matrices with environment-aware skip sets, enabling similar efficiency gains for robot–environment collision queries.

\subsection{Broader Impact}
By making the generation, inspection, and refinement of self-collision matrices more efficient and transparent, our system has the potential to accelerate many aspects of robotics research and development. Faster and more accurate collision checking directly translates to quicker planning, more robust simulation, and safer real-world execution. Importantly, the tool also provides an educational benefit: students and practitioners can interactively explore how robot geometry and kinematics interact, deepening their intuition for collision behavior. 

\subsection{Summary}
In summary, our tool provides a new lens on a long-standing problem in robotics infrastructure. Through automatic generation, multiple shape representations, dynamic visualization, and compatibility with existing baselines, the system closes the gap between approximation and reliability. While further work is needed to extend scalability and automation, the results of this study demonstrate that interactive, shape-type–aware self-collision matrices represent a practical and effective step forward for the robotics community.





\bibliographystyle{plainnat}
\bibliography{refs}


\end{document}